\documentclass[runningheads]{llncs}

\usepackage{eccv}

\usepackage{eccvabbrv}
\usepackage{graphicx}
\usepackage{booktabs}
\usepackage{array}
\usepackage{amsmath,amssymb}
\usepackage{tikz}
\usetikzlibrary{arrows.meta,positioning}
\usepackage[accsupp]{axessibility}

\usepackage[hidelinks]{hyperref}
\hypersetup{
  pdftitle={Trajectory Design and Budgeted Querying for Digital Twin Calibration},
  pdfauthor={Vladyslava Spitkovska and Dmytro Kuzmenko},
  pdfkeywords={data curation, active learning, reinforcement learning, system identification, digital twins}
}

\graphicspath{{figures/}}

\definecolor{cstage}{RGB}{65,140,225}
\definecolor{cstageborder}{RGB}{35,105,185}
\definecolor{coracle}{RGB}{240,155,60}
\definecolor{coracleborder}{RGB}{215,125,35}
\definecolor{cflow}{RGB}{90,90,90}
\definecolor{clbl}{RGB}{60,60,60}

\begin{document}

\title{Trajectory Design and Budgeted Querying for Digital Twin Calibration}
\titlerunning{Trajectory Design and Budgeted Querying for DT Calibration}

\author{
Vladyslava Spitkovska\inst{1} \and
Dmytro Kuzmenko\inst{2,3}
}

\authorrunning{V. Spitkovska and D. Kuzmenko}

\institute{
Department of Mathematics,\\
National University of Kyiv-Mohyla Academy, Kyiv, Ukraine\\
\email{v.spitkovska@ukma.edu.ua}
\and
Department of Multimedia Systems,\\
National University of Kyiv-Mohyla Academy, Kyiv, Ukraine\\
\email{kuzmenko@ukma.edu.ua}
\and
Department of Computer Science,\\
University of Turin, Turin, Italy\\
\email{dmytro.kuzmenko@unito.it}
}

\maketitle

\begin{abstract}
Digital-twin calibration requires interaction data that is expensive to collect. We study two acquisition decisions: which trajectories to generate, and when to spend a limited budget on privileged parameter measurements. Our framework couples an excitation-oriented reinforcement learning controller, a recurrent parameter estimator with predictive uncertainty, and a budgeted query policy. In Pendulum, a Random Forest diagnostic recovers gravity only weakly from task-oriented trajectories and does not recover mass or length, while a GRU trained on excitation-oriented trajectories reaches a mean absolute error of $0.0066$ with no queries. We then withdraw continuous oracle access partway through an episode, so that the twin must run on the estimator's output for the remainder. The estimator-plus-policy pipeline achieves a terminal error of $0.0092$ under a three-query budget, against $0.2031$ for an uncalibrated twin. In partially observable Waterworld, five controllers produce different observed error profiles across three hidden parameters, and an estimator trained on a five-controller mixture reaches online normalized errors of roughly 4--5\%. These exploratory case studies are not controlled ablations, but they motivate treating trajectory design and query allocation as explicit design variables in data-scarce calibration.
\end{abstract}

\section{Introduction}
\label{sec:intro}

Data-centric machine learning shows that which data a model sees matters as much as the model itself: pruning, distillation, and active selection trade a costly, unconstrained dataset for a smaller, deliberately assembled one \cite{settles2009active,sorscher2022pruning,wang2018dataset}. The trade-off is sharpest where each example requires querying an expensive or safety-critical source. Digital twins (DTs) exemplify the problem: a DT is useful only while its parameters track the real system, and closing that gap requires interaction data that is rarely free \cite{tian2022realtime,lambert2020digitaltwin}.

A common calibration pipeline separates data collection from parameter fitting, collecting a fixed dataset under an unrelated control objective before fitting the parameters. We instead study calibration through two coupled acquisition decisions: which trajectories to generate, and when a privileged parameter measurement is worth its cost under a budget. The framework combines established components within a data-centric formulation that connects trajectory design to measurement allocation, together with diagnostics that characterize each decision.

We formulate calibration around these decisions and combine an excitation-oriented controller, a recurrent estimator with predictive uncertainty, and a budgeted query policy. We report exploratory evidence in three regimes: recoverability diagnostics in Pendulum, calibration after continuous oracle access is withdrawn mid-episode, and controller-conditioned error profiles in partially observable Waterworld. We train the components in sequence rather than jointly: the controller first, the estimator on its trajectories, and the query policy against a frozen estimator. The Pendulum and Waterworld studies therefore exercise different parts of the framework rather than one integrated pipeline.

\section{Related Work}
\label{sec:related}
Optimal experimental design and Bayesian experimental design formalize which data to collect next in order to maximize information about unknown parameters \cite{chaloner1995bayesian}, and dual control extends this to closed-loop control under parameter uncertainty \cite{mesbah2018stochastic}. Receding Horizon Curiosity makes the sequential version explicit, interleaving episodic exploration with Bayesian nonlinear system identification \cite{schultheis2020rhc}. We view our excitation reward and query policy as a learned instance of the same classical idea, with a hand-designed rather than information-theoretic objective. Active learning selects points from a fixed pool \cite{settles2009active}, while pruning and distillation compress an existing dataset \cite{sorscher2022pruning,wang2018dataset}; we act upstream of both, since our controller generates the trajectories rather than selecting from a pool.

RL-driven excitation for system identification has targeted sim-to-real transfer \cite{memmel2024asid,lambert2020digitaltwin,chebotar2019closing} and active domain randomization \cite{mehta2020active}. Prior work has also studied active learning of dynamical systems outside RL \cite{wagenmaker2020active}, while simulation-based inference addresses the related inverse problem of fitting simulator parameters to observed data \cite{cranmer2020frontier}. Closest to our setting, SPI-Active identifies physical parameters of legged robots by sampling-based optimization and improves data informativeness by optimizing an exploration policy's commands to maximize the Fisher information of the collected trajectories \cite{sobanbabu2025spiactive}. That work derives its excitation signal from an explicit information objective. In contrast, we use a hand-designed squared sensitivity proxy and focus on the complementary decision of when to request a costly measurement under a budget. The excitation mechanism follows established active system-identification practice; our framework couples it with a recurrent uncertainty-aware estimator \cite{kendall2017uncertainties} and a budgeted query policy. We model calibration as a POMDP \cite{spaan2012pomdp} with PPO \cite{schulman2017ppo} controllers and a GRU \cite{chung2014gru} history encoder.

\section{Method}
\label{sec:method}

\subsection{Problem formulation}
\label{sec:formulation}
The environment is a POMDP $\mathcal{M}=(\mathcal{S},\mathcal{A},\mathcal{O},T_\vartheta,R_\vartheta)$ with hidden parameters $\vartheta\in\Theta$ that are constant within an episode. Calibration seeks an estimate of $\vartheta$ from the interaction history $\tau_t=(o_1,a_1,r_1,\dots,o_t,a_t,r_t)$, subject to $Q\le B$, where $q_t\in\{0,1\}$ indicates a query at step $t$, $Q=\sum_t q_t$, and $B$ is a hard budget. A query returns the environment's current parameter vector $\vartheta_{\mathrm{true}}$ at a constant cost $c_{\mathrm{query}}$. This models privileged access to the ground-truth state of a simulator or twin, analogous to an expensive sensor readout or laboratory measurement. Note that $\vartheta$ is hidden from the estimator even where the dynamical state is fully observed, as in Pendulum: full state observability does not make the parameter directly readable.

\subsection{General framework}
\label{sec:framework}
We instantiate the formulation with three components: a controller that generates the trajectories, a recurrent estimator that turns them into a parameter estimate with predictive uncertainty, and a query policy that decides when a measurement is worth its cost.

\paragraph{Controller.} A controller trained for nominal task reward need not generate trajectories that are informative about $\vartheta$: it may drive the system toward states from which the hidden parameters are barely observable, or suppress exactly the motion that would expose them. We therefore treat the trajectory-generation objective as a design decision and shape the controller reward toward \emph{excitation} of the parameters of interest, keeping the shaped terms interpretable rather than deriving an information-theoretic objective. In Waterworld a single excitation reward was not sufficient, and we instead assemble a mixture of independently shaped controllers ($\S$\ref{sec:water}).

\paragraph{Estimator.} A GRU $f_\psi$ maps the trajectory prefix to a predictive mean and standard deviation, $\big(\hat\vartheta(t),\sigma(t)\big)=f_\psi(o_{1:t},a_{1:t})$ with $h_t=\mathrm{GRU}(h_{t-1},x_t)$, where $x_t$ is a per-step feature vector built from the observation, action, and reward. At every step a feed-forward head reads out $h_t$ and emits $\hat\vartheta(t)$ and $\log\sigma(t)$, which parameterize a diagonal Gaussian over $\vartheta$. In Pendulum, $x_t$ concatenates eight engineered features derived from the observation, action, and reward. In Waterworld, we reduce the $242$-dimensional observation to $47$ engineered features before concatenating the action and reward (Table~\ref{tab:impl}). We train the GRU first with MSE and then with a Gaussian negative log-likelihood, so $\sigma(t)$ is fitted rather than fixed. The recurrent state is a deterministic history embedding that parameterizes a predictive Gaussian, not a Bayesian posterior.

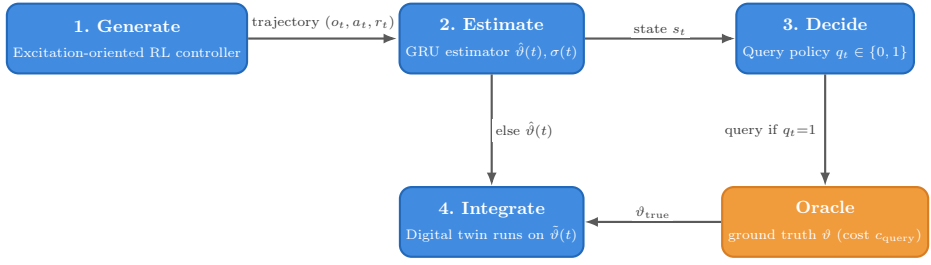
\begin{figure}[!tb]
\centering
\resizebox{\linewidth}{!}{%
\begin{tikzpicture}[
  font=\small, >=Latex, node distance=2.0cm and 2.7cm,
  stage/.style={rounded corners=6pt, minimum width=2.8cm, minimum height=1.25cm, align=center,
                draw=cstageborder, line width=1.1pt,
                fill=cstage, text=white},
  oracle/.style={rounded corners=6pt, minimum width=2.8cm, minimum height=1.25cm, align=center,
                 draw=coracleborder, line width=1.1pt,
                 fill=coracle, text=white},
  flow/.style={draw=cflow, line width=1.1pt, -{Latex[length=2.4mm]}},
  lbl/.style={font=\scriptsize, text=clbl, align=center,
              fill=white, inner sep=1.5pt}
]
\node[stage] (ctrl) {\textbf{1. Generate}\\[2pt]{\scriptsize\color{white!85}Excitation-oriented RL controller}};
\node[stage, right=of ctrl] (gru) {\textbf{2. Estimate}\\[2pt]{\scriptsize\color{white!85}GRU estimator $\hat\vartheta(t),\sigma(t)$}};
\node[stage, right=of gru] (calib) {\textbf{3. Decide}\\[2pt]{\scriptsize\color{white!85}Query policy $q_t\in\{0,1\}$}};
\node[stage, below=of gru] (twin) {\textbf{4. Integrate}\\[2pt]{\scriptsize\color{white!85}Digital twin runs on $\tilde\vartheta(t)$}};
\node[oracle, below=of calib] (oracle) {\textbf{Oracle}\\[2pt]{\scriptsize\color{white!90}ground truth $\vartheta$ (cost $c_{\mathrm{query}}$)}};

\draw[flow, shorten >=0.8pt] (ctrl) -- node[lbl, above, yshift=2.5pt] {trajectory $(o_t,a_t,r_t)$} (gru);
\draw[flow] (gru) -- node[lbl, above] {state $s_t$} (calib);
\draw[flow] (calib.south) -- node[font=\scriptsize, text=clbl,
  anchor=east, inner sep=0pt, pos=0.5, xshift=-4pt] {query if $q_t{=}1$} (oracle.north);
\draw[flow] (oracle) -- node[lbl, above] {$\vartheta_{\mathrm{true}}$} (twin);
\draw[flow] (gru.south) -- node[lbl, right] {else $\hat\vartheta(t)$} (twin.north);
\end{tikzpicture}%
}
\caption{The acquisition stages. The controller generates a trajectory (stage 1); the GRU turns it into an estimate with predictive uncertainty (stage 2); the query policy decides whether to issue a measurement query (stage 3); the twin integrates the deployed parameter $\tilde\vartheta(t)$ defined in \cref{eq:deployed} (stage 4). The controller reads environment observations only, so there is no feedback edge from the twin back to the controller.}
\label{fig:pipeline}
\end{figure}

\paragraph{Deployed parameter and query policy.} We distinguish the estimator's raw output $\hat\vartheta(t)$ from the parameter $\tilde\vartheta(t)$ actually supplied to the twin. The twin integrates $\tilde\vartheta(t)$, and
\begin{equation}
\tilde\vartheta(t) = \begin{cases}
\vartheta_{\mathrm{true}} & \text{if } q_t=1 \text{ and } Q_{t-1}<B,\\
\hat\vartheta(t) & \text{otherwise.}
\end{cases}
\label{eq:deployed}
\end{equation}
The oracle value is not among the estimator's input features, so $\hat\vartheta$ is unaffected by a query and $\tilde\vartheta(t{+}1)$ returns to the estimator's own output at the next step. A query is therefore a one-step correction of the deployed parameter, not a correction of the estimator. The controller acts on environment observations and never conditions on $\tilde\vartheta$, so a query does not alter the trajectory being generated either. The query policy $\pi_{\mathrm{q}}$ observes a compact state derived from episode progress, the estimator's predictive uncertainty and standardized output, and the remaining budget; we train it with PPO against a frozen estimator with a per-step query cost plus a terminal accuracy term. The policy-specific state and reward for the Pendulum instance are given in \cref{sec:pendulum-inst}. \Cref{fig:pipeline} summarizes the four acquisition stages and the separation between the estimator output and the deployed parameter.

We report the terminal error $\mathrm{MAE}=\mathbb{E}\|\tilde\vartheta(T)-\vartheta\|$ on the deployed parameter, normalized by each parameter's range $\vartheta_{\max}-\vartheta_{\min}$ where useful, together with the realized query count.

\subsection{Pendulum instantiation}
\label{sec:pendulum-inst}
Pendulum is a fully observed testbed with a single hidden parameter. The state is the pole angle $\phi_t$, measured from the upright position, and its angular velocity $\omega_t$; the observation $o_t=(\cos\phi_t,\sin\phi_t,\omega_t)\in\mathbb{R}^3$ is fully observable up to angular periodicity. The hidden gravity $g$ enters the dynamics only through
\begin{equation}
\ddot\phi_t = -\frac{3g}{2l}\,\sin\phi_t + \frac{3}{ml^2}\,u_t,
\label{eq:pendulum-dyn}
\end{equation}
with applied torque $u_t\in[-2,2]$ and fixed mass $m$ and length $l$. The sensitivity of the dynamics to $g$ is thus proportional to $\sin\phi_t$, and a controller rewarded for stabilizing near $\phi_t\approx0$ suppresses its own identification signal. We therefore shape the controller reward toward excitation,
\begin{equation}
r_t^{\mathrm{ctrl}} = \sin^2(\phi_t) - \lambda_a u_t^2 - \lambda_\omega\,\mathrm{ReLU}(|\omega_t|-\omega_{\max})^2,
\label{eq:excitation}
\end{equation}
where $\lambda_a,\lambda_\omega$ weight control effort and an angular-speed cap $\omega_{\max}$. The leading term is a nonnegative \emph{squared sensitivity proxy}: it is largest where the dynamics respond most strongly to $g$ and vanishes at the upright equilibrium a task-optimal controller drives toward. We use it as a proxy and do not derive the Fisher information of $g$ for this system.

For this instance the query policy observes the four-dimensional state
\begin{equation}
s_t=\big(\;t/T,\;\;10\,\sigma(t),\;\;(\hat\vartheta(t)-\mu)/\varsigma,\;\;Q_t/B\;\big),
\label{eq:qstate}
\end{equation}
where $t/T$ is episode progress, $\sigma(t)$ is rescaled for numerical conditioning, $\hat\vartheta(t)$ is standardized by the training-set mean $\mu$ and standard deviation $\varsigma$, and $Q_t=\sum_{s\le t}q_s$ is the budget consumed so far. It is a two-hidden-layer MLP with 64 units per layer, which we train with PPO against a frozen estimator under
\begin{equation}
r_t^{\mathrm{q}} = -c_{\mathrm{query}}\,q_t,\qquad
R_{\mathrm{terminal}} = -\gamma\,\|\tilde\vartheta(T)-\vartheta\|,\qquad \text{s.t. } Q\le B,
\label{eq:qreward}
\end{equation}
with $c_{\mathrm{query}}=1.0$, $\gamma=5.0$, and $B=3$. The only action-dependent term in the per-step reward is the query cost, so every incentive to query comes from the terminal accuracy term: the policy queries when it expects the resulting deployed value to offset its cost by the end of the episode. This state and objective are specific to Pendulum. We do not train or evaluate a query policy in Waterworld, where only the controller mixture and the estimator are studied.

\subsection{Waterworld instantiation}
\label{sec:waterworld-inst}
Waterworld is a partially observable, single-agent environment in which a pursuer moves among food and poison objects. The action is a two-dimensional acceleration $a_t\in\mathbb{R}^2$, which we normalize when its norm exceeds the hidden maximum acceleration $\mathrm{pursuer\_max\_accel}$; the velocity integrates as $v_{t+1}=v_t+a_t$. We build the observation $o_t\in\mathbb{R}^{242}$ from sensor readings of distances to, and interactions with, nearby objects; the hidden sensor range directly shapes this observation function, so the parameters influence what the agent can perceive. We randomize and calibrate three parameters per episode: sensor range in $[0.20,0.35]$, maximum acceleration in $[0.35,0.70]$, and pursuer speed in $[0.12,0.35]$ (Table~\ref{tab:impl}). We describe the controller mixture used to generate trajectories in this environment and its evaluation in \cref{sec:water}.

\section{Experiments}
\label{sec:exp}
We train controllers with Stable-Baselines3 \cite{raffin2021sb3}, implement the estimator and query policy in PyTorch, and serve the oracle over a REST API. Table~\ref{tab:impl} lists the configuration.

We use two environments. Pendulum is fully observed and has a single hidden parameter $g$; Waterworld is partially observable, with $\mathrm{dim}(\vartheta){=}3$ and $o_t\in\mathbb{R}^{242}$. Unless otherwise stated, we add uniform noise to Pendulum parameter values returned by the oracle; the withdrawal protocol in \cref{sec:withdrawal} uses exact oracle values. Environment observations and Waterworld oracle values are noiseless. All reported spreads are standard deviations across evaluation episodes, from a single training run per configuration, so we report no significance tests.

\begin{table}[!tb]
\caption{Implementation details.}
\label{tab:impl}
\centering
\footnotesize
\begin{tabular}{@{}>{\raggedright\arraybackslash}p{2.4cm}>{\raggedright\arraybackslash}p{4.0cm}>{\raggedright\arraybackslash}p{4.7cm}@{}}
\toprule
 & Pendulum & Waterworld \\
\midrule
Hidden parameters & gravity $g$ & sensor range $[0.20,0.35]$, maximum acceleration $[0.35,0.70]$, pursuer speed $[0.12,0.35]$ \\
Episode length & $T{=}200$; $T{=}300$ in \cref{sec:withdrawal} & $T{=}500$ \\
Training episodes & 500, 80/20 split at episode level & 400--550 per estimator version \\
Validation & 80/20 episode-level hold-out & 5-fold cross-validation, or a stratified 25\% hold-out for the final version \\
Evaluation & 20 episodes & 400 online episodes across five policies (Table~\ref{tab:mixture}) \\
Controller & PPO, 100k steps, 4 parallel environments & PPO \\
Estimator input & 8 engineered features & 47 engineered features from the 242-dim observation \\
GRU & hidden 128, 2 layers, dropout 0.15, head $128{\to}64{\to}32{\to}(\mu,\log\sigma)$ & hidden 192 \\
Objective & MSE for 80 epochs, then Gaussian NLL & MSE, then Gaussian NLL \\
Optimization & Adam $3{\times}10^{-4}$, ReduceLROnPlateau (patience 15), gradient clip 1.0, batch 32, ${\le}150$ epochs & batch 32 \\
Query policy & 4-dim state (\cref{eq:qstate}), MLP $2{\times}64$, PPO, ${\approx}300$ live-oracle episodes, estimator frozen & none \\
Budget and cost & $B{=}3$, $c_{\mathrm{query}}{=}1.0$, $\gamma{=}5.0$ & none \\
Oracle noise & $\varepsilon\sim\mathcal{U}(-\delta,\delta)$ on the returned $g$; none in \cref{sec:withdrawal} & none \\
\bottomrule
\end{tabular}
\end{table}

\subsection{Recoverability diagnostics in Pendulum}
\label{sec:pendulum}
We ran two diagnostics. First, we trained standard stabilizing controllers and fitted a Random Forest that predicts each physical parameter from episode-summary statistics of their trajectories. Table~\ref{tab:rf} reports the result: gravity is only weakly predictable ($R^2\approx0.02$), and mass and length are not predicted better than the training mean. This diagnostic measures recoverability only for the chosen estimator and trajectory distribution. It does not establish formal non-identifiability, and other features or estimators may recover information it misses.

\begin{table}[!tb]
\caption{Random Forest diagnostic on trajectories from task-oriented stabilizing controllers, predicting each physical parameter from episode-summary statistics.}
\label{tab:rf}
\centering
\begin{tabular}{@{}lcc@{}}
\toprule
Parameter & MAE & $R^2$ \\
\midrule
$g$ (gravity) & 0.311 & 0.019 \\
$m$ (mass) & 0.051 & $-0.005$ \\
$l$ (length) & 0.045 & $-0.121$ \\
\bottomrule
\end{tabular}
\end{table}

Second, we trained the GRU estimator on trajectories from the excitation-shaped controller of \cref{eq:excitation} and evaluated it on a static, episode-constant $g$ (Table~\ref{tab:pendulum}). It reaches $\mathrm{MAE}(g)=0.0066$ with no queries, and the budgeted query policy reaches $0.0080$ while spending $0.20$ queries per episode on average. At this sample size, the difference does not support a ranking. Because the diagnostics use different estimators and data distributions, their comparison does not isolate the effect of trajectory design; doing so requires a matched estimator and episode count.

\begin{table}[!tb]
\caption{Pendulum with a static $g$ per episode, over 20 evaluation episodes (mean $\pm$ standard deviation).}
\label{tab:pendulum}
\centering
\begin{tabular}{@{}lcc@{}}
\toprule
Strategy & $\mathrm{MAE}(g)$ & Queries/ep. \\
\midrule
GRU estimator, no queries & $0.0066\pm0.0053$ & 0.00 \\
GRU estimator + budgeted query policy & $0.0080\pm0.0067$ & 0.20 \\
\bottomrule
\end{tabular}
\end{table}

\subsection{Calibration after withdrawal of continuous oracle access}
\label{sec:withdrawal}
We next place the estimator and query policy of \cref{sec:pendulum}, without retraining, in the twin-calibration loop. For the first 50 steps the twin receives the ground-truth parameter directly from the oracle, so the twin and oracle trajectories coincide. From step 51 we withdraw that access: the twin must run on $\tilde\vartheta(t)$ from \cref{eq:deployed} for the remaining 250 steps. We hold $\tilde\vartheta$ fixed over the final 100 steps, so the terminal metric $|\tilde\vartheta(T)-\vartheta|$ scores a single committed value. The ground-truth parameter does not change during the episode, and its value lies inside the range seen in training; what is new is the withdrawal of access, not the parameter.

\begin{figure}[!tb]
\centering
\includegraphics[width=\linewidth]{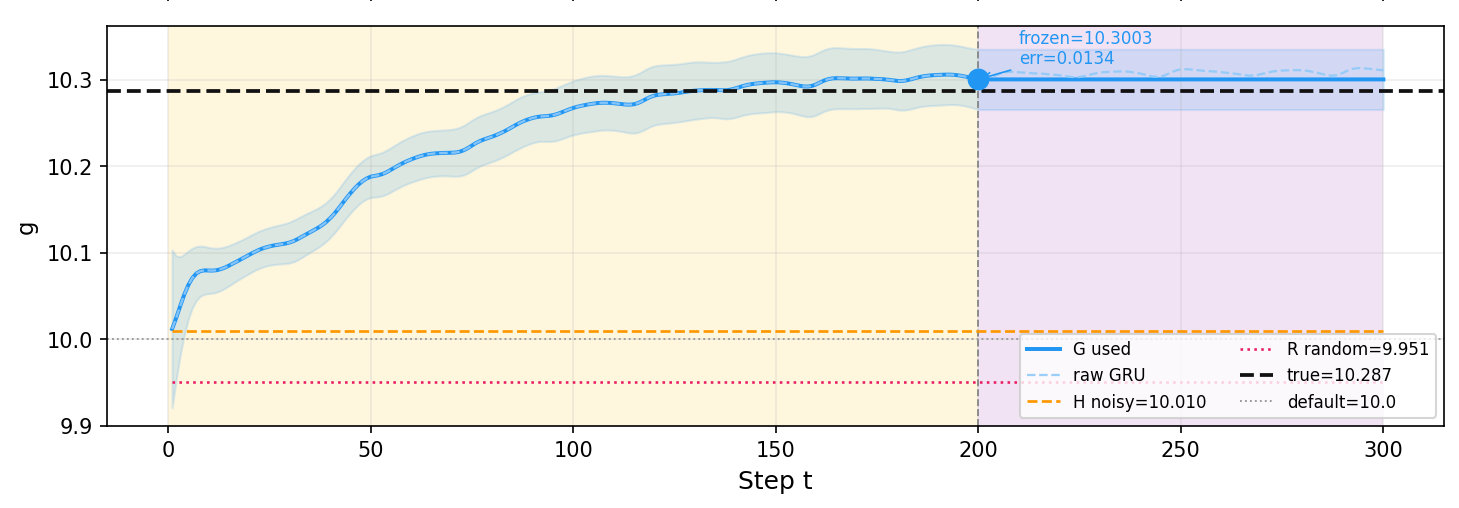}
\caption{One episode with $g_{\mathrm{true}}=10.287$ (black dashed), in which no query was triggered. The deployed value $\tilde\vartheta(t)$ (solid) and the raw estimator output $\hat\vartheta(t)$ (light dashed) are shown with the estimator's $\pm\sigma(t)$ band. Shading marks the online phase (steps 1--200) and the frozen phase (201--300), where the committed value $10.3003$ gives a terminal error of $0.0134$. Horizontal lines mark the uncalibrated default ($10.0$) and two constant reference values.}
\label{fig:conv}
\end{figure}

Table~\ref{tab:withdrawal} reports the terminal error over 20 episodes. The estimator-plus-policy pipeline ends at $0.0092$ within a hard budget of three queries, against $0.2031$ for a twin that holds its uncalibrated default value and $0.3047$ for a randomly drawn parameter. This comparison evaluates the combined pipeline but not the query policy's marginal contribution. Under the static protocol in Table~\ref{tab:pendulum}, the estimator alone reaches $0.0066$, suggesting that much of the accuracy may not require querying. A query-free control under the same 300-step protocol is needed to separate these effects; the no-query episode in Figure~\ref{fig:conv} is illustrative rather than a substitute for that comparison. We omit the heuristic query baselines because they used noisy oracle readings and piecewise-constant updates, whereas this protocol uses exact readings and continuous updates. Their differences would therefore extend beyond query timing.

\begin{table}[!tb]
\caption{Terminal error on the deployed parameter after continuous oracle access is withdrawn at step 51, over 20 episodes.}
\label{tab:withdrawal}
\centering
\begin{tabular}{@{}lcc@{}}
\toprule
Deployed parameter & $|\tilde\vartheta(T)-\vartheta|$ & Query budget \\
\midrule
Continuous oracle access (reference) & 0.0000 & N/A \\
GRU estimator + budgeted query policy & 0.0092 & $\le 3$ \\
Uncalibrated default value & 0.2031 & 0 \\
Randomly drawn parameter & 0.3047 & 0 \\
\bottomrule
\end{tabular}
\end{table}

Figure~\ref{fig:conv} traces the estimator over one episode. The estimate rises from the uncalibrated default toward the true value over roughly 130 steps, while its predicted uncertainty contracts. The protocol fixes the committed value for the last 100 steps, and oracle measurements never enter the estimator's inputs; a query therefore corrects the deployed parameter for one step without shifting the subsequent estimate. Because we did not log realized query times, the figure reports estimator convergence only.

\subsection{Controller-conditioned error profiles in Waterworld}
\label{sec:water}
Waterworld has three hidden parameters and a 242-dimensional observation, and a single excitation reward did not suffice here. We assembled a non-uniform pool of five controllers: four PPO controllers ($\pi_1$ broad-exploration, $\pi_2$ velocity-conservative, $\pi_3$ trained under randomized $\vartheta$, and $\pi_4$ acceleration- and diversity-focused) plus a deterministic zigzag policy. We chose the mixture weights by downstream recovery rather than episode reward. Higher task reward did not track more informative trajectories: one retrained controller raised mean episode reward by $4.5\%$ but degraded parameter recovery, so we dropped it in favour of its predecessor; we kept another despite a $4.4\%$ reward decrease because it produced $42.1\%$ more food-contact events and more diverse velocity profiles.

Each controller contributes a fixed share of the estimator's training trajectories and generates a disjoint block of online evaluation episodes, as Table~\ref{tab:mixture} shows. The evaluation therefore measures generalization across parameter draws and episodes, not to controller families held out from training. The unequal block sizes also warrant caution, particularly the $n{=}20$ block for $\pi_1$.

\begin{table}[!tb]
\caption{Final estimator, evaluated online: normalized error per hidden parameter, by the controller that generated the evaluation episodes. ``Share'' is that controller's proportion of the training mixture, $n$ its number of evaluation episodes. Bold marks the lowest observed value per column.}
\label{tab:mixture}
\centering
\small
\begin{tabular}{@{}lrrcccc@{}}
\toprule
Controller & Share & $n$ & Sensor range & Max.\ acceleration & Pursuer speed & Mean \\
\midrule
$\pi_1$ & 5\% & 20 & 0.0384 & 0.0458 & 0.0689 & 0.0510 \\
$\pi_2$ & 10\% & 60 & 0.0368 & \textbf{0.0300} & 0.0703 & 0.0457 \\
$\pi_3$ & 40\% & 112 & 0.0327 & 0.0409 & \textbf{0.0513} & \textbf{0.0416} \\
$\pi_4$ & 35\% & 148 & \textbf{0.0326} & 0.0370 & 0.0603 & 0.0433 \\
zigzag & 10\% & 60 & 0.0367 & 0.0418 & 0.0608 & 0.0464 \\
\bottomrule
\end{tabular}
\end{table}

The controllers produce different observed error profiles rather than exclusive coverage: $\pi_4$ gives the lowest error for sensor range, $\pi_2$ for maximum acceleration, and $\pi_3$ for pursuer speed and on the mean. These profiles motivate a mixture but do not show that any parameter is recoverable only under one regime. Errors span $3$--$7\%$ of each parameter's range, and pursuer speed is the most difficult to recover, with the largest error in every row and a worst case of $0.0703$.

An unmatched checkpoint comparison also illustrates the offline-to-online gap. The checkpoint selected by offline validation attained the lowest offline value we measured ($0.0373$ mean normalized error) but degraded to $0.1024$ online on a later batch of episodes. The final estimator reached $0.0416$--$0.0510$ online, the range of per-controller means in Table~\ref{tab:mixture}. Because the checkpoints differ in training mixture, architecture (including separate uncertainty heads per parameter and a hidden-size reduction from 224 to 192), training-data volume, and evaluation protocol, this comparison cannot attribute the online difference to mixture rebalancing. It shows only that offline selection error did not predict online error in these runs.

\section{Discussion and Limitations}
\label{sec:discussion}
Across the experiments, calibration outcomes vary with both the estimator and the data-acquisition process. In Pendulum, query-free accuracy on excitation-oriented trajectories leaves little measurable benefit for the query policy in the static setting (\cref{sec:pendulum}). After continuous oracle access ends, the combined pipeline retains low terminal error, but the unmatched protocol does not isolate querying (\cref{sec:withdrawal}). In Waterworld, controller-conditioned errors vary across observed trajectory distributions, and offline validation did not predict later online error (\cref{sec:water}). These case studies establish feasibility and sensitivity to data collection and evaluation, not causal improvements over standard identification methods.

We trained the components stagewise and ran each configuration once. The protocols also differ in estimator choice, data volume, architecture, noise, update rules, and evaluation episodes. These differences prevent causal attribution to trajectory design, querying, or mixture rebalancing. In the static Pendulum setting, the policy used $0.20$ queries per episode under $B{=}3$, so the budget did not bind. We did not evaluate uncertainty coverage or record query times. Both environments are simulated and have low-dimensional hidden parameters, so scalability to more complex systems remains open.

The most informative next comparison is a query-free estimator under the same 300-step withdrawal protocol. Matched trajectory comparisons, classical filters or least-squares baselines, and a binding-budget regime would isolate the remaining contributions. Feeding queried measurements into the recurrent state would make corrections persistent, while an information-theoretic objective could replace the hand-designed excitation proxy \cite{sobanbabu2025spiactive,schultheis2020rhc}. Evaluation on higher-dimensional and physical systems would then test scalability.

\bibliographystyle{splncs04}
\bibliography{main}

\end{document}